\documentclass{llncs}

\usepackage{array}
\usepackage{graphicx}
\usepackage{hyperref}
\usepackage{xcolor}
\usepackage{listings}

\definecolor{codegreen}{rgb}{0,0.6,0}
\definecolor{codegray}{rgb}{0.5,0.5,0.5}
\definecolor{codepurple}{rgb}{0.58,0,0.82}
\definecolor{backcolour}{rgb}{0.97,0.97,0.97}

\lstdefinestyle{python_jay}{
    backgroundcolor=\color{backcolour},
    commentstyle=\color{codegreen},
    keywordstyle=\color{blue},
    numberstyle=\tiny\color{codegray},
    stringstyle=\color{codepurple},
    basicstyle=\ttfamily\footnotesize,
    breakatwhitespace=false,
    breaklines=true,
    captionpos=b,
    keepspaces=true,
    numbers=left,
    numbersep=5pt,
    showspaces=false,
    showstringspaces=false,
    showtabs=false,
    tabsize=2
}

\title{The FreshPRINCE: A Simple Transformation Based Pipeline Time Series Classifier}
\author{Matthew Middlehurst and Anthony Bagnall}

\institute{
School of Computing Sciences, University of East Anglia, UK\\
\email{M.Middlehurst@uea.ac.uk}
}

\begin{document}

\maketitle

\begin{abstract}

There have recently been significant advances in the accuracy of algorithms proposed for time series classification (TSC). However, a commonly asked question by real world practitioners and data scientists less familiar with the research topic, is whether the complexity of the algorithms considered state of the art is really necessary. Many times the first approach suggested is a simple pipeline of summary statistics or other time series feature extraction approaches such as TSFresh, which in itself is a sensible question; in publications on TSC algorithms generalised for multiple problem types, we rarely see these approaches considered or compared against. We experiment with basic feature extractors using vector based classifiers shown to be effective with continuous attributes in current state-of-the-art time series classifiers. We test these approaches on the UCR time series dataset archive, looking to see if TSC literature has overlooked the effectiveness of these approaches. We find that a pipeline of TSFresh followed by a rotation forest classifier, which we name FreshPRINCE, performs best. It is not state of the art, but it is significantly more accurate than nearest neighbour with dynamic time warping, and represents a reasonable benchmark for future comparison.


\end{abstract}

\section{Introduction}

A wide range of complex algorithms for time series classification (TSC) have been proposed. These include ensembles of deep neural networks~\cite{fawaz20inception}, heterogeneous meta-ensembles build on different representations~\cite{middlehurst21hc2}, homogeneous ensembles with embedded representations~\cite{shifaz20ts-chief} and randomised kernels~\cite{dempster20rocket}. The majority of these algorithms rely on some form of transformation: features that in some way model the discriminatory time characteristics are extracted and used in the classification process. These features are often very complex, and usually embedded in the classifiers in complicated ways. For example, the Temporal Dictionary Ensemble (TDE)~\cite{middlehurst20temporal} is centred around the Symbolic Fourier Approximation (SFA)~\cite{schafer12sfa} transformation. The transform itself simply discretises the series into a set of words using a sliding window. However, just performing the transform does not lead to an algorithm that is competitive in accuracy. TDE also employs a spacial pyramid, uses bi-gram frequency, a bespoke distance function and a Gaussian process based parameter setting mechanism. The complexity increases further if the data is multivariate, containing multiple time series per case.

\begin{figure}[t]
    	\centering
        \includegraphics[width=\linewidth,trim={0cm 0cm 0cm 0cm},clip]{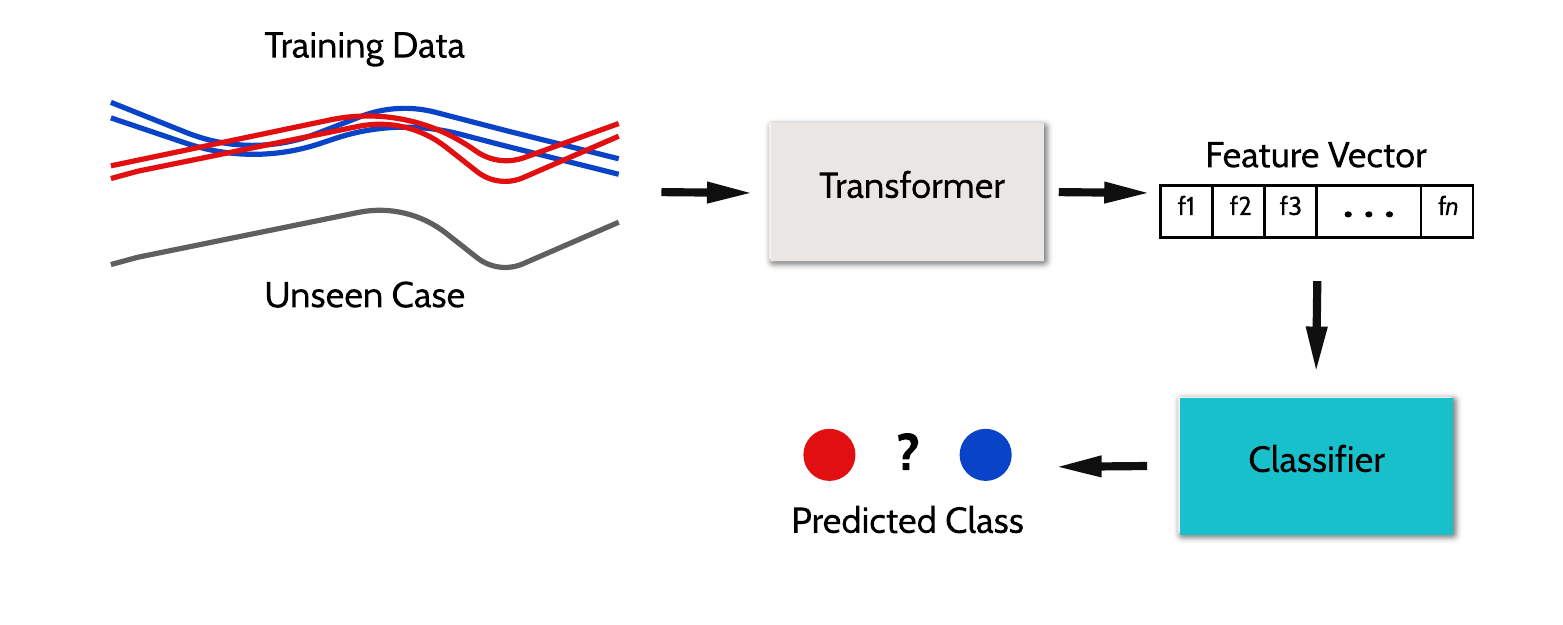}
        \caption{Visualisation of a simple pipeline algorithm for TSC. Could using standard transformers and vector based classifiers be as good as state of the art TSC algorithms?}
        \label{fig:pipeline}
    \end{figure}

Researchers not directly involved in TSC algorithm research, and data scientists in particular, often ask the not unreasonable question of whether these complicated representations are really necessary to get a good classifier. They wonder whether a simple pipeline using standard feature extractors, as illustrated in Figure~\ref{fig:pipeline} would not in fact be at least as good as complicated classifiers claiming to be state of the art? Clearly, the answer will not be the same for all problems, and the detailed answer depends on what level of accuracy is deemed sufficient for a particular application. However, we can address the hypothesis of whether, on average, a standard pipeline of transformation plus classifier performs as well as bespoke benchmarks and state of the art.
 Specifically, we compare a range of pipeline combinations of off the shelf unsupervised time series transformers with commonly used vector based classifiers to the current state of the art in TSC as described in~\cite{middlehurst21hc2}. In Section~\ref{sec:background} we describe the transformers and classifiers used in our pipeline experiments, and give a brief overview of the state of the art in TSC. In Section~\ref{sec:experiments} we describe our experimental structure, and in Section~\ref{sec:results} we present our findings. Finally, in Section~\ref{sec:conclusions} we draw our conclusions and summarise what we have learnt from this study.

\section{Background}
\label{sec:background}

TSC algorithms tend to follow one of three structures. the simplest involves single pipelines such as that described in Section~\ref{fig:pipeline}, where the transformation is either supervised (e.g. Shapelet Transform Classifier~\cite{bostrom17binary}) or unsupervised (e.g. ROCKET~\cite{dempster20rocket}). These algorithms tend to involve an over-produce and select strategy: a huge number of features are created, and the classifier is left to determine which are most useful. The transform can remove time dependency, e.g. by calculating summary features. We call this type series-to-vector transformations. Alternatively, they may be series-to-series, transforming into an alternative time series representation where we hope the task becomes more easily tractable (e.g. transforming to the frequency domain of the series).

The second transformation based design pattern involves ensembles of pipelines, where each base pipeline consists of making repeated, different, transforms and using a homogeneous base classifier (e.g. Canonical Interval Forest~\cite{middlehurst20canonical}). These ensembles can also be heterogeneous, collating the classifications from transformation pipelines and ensembles of differing representations of the time series (e.g. HIVE-COTE~\cite{middlehurst21hc2}).

The third common pattern involves transformations embedded inside a classifier structure. An example of this is a decision tree: where the data is transformed, or a distance measure is applied prior to any splitting criteria at each node (e.g. TS-CHIEF~\cite{shifaz20ts-chief}).

\subsection{State of the art for TSC}

The state-of-the art for TSC consists of one classifier from each of the structures described, as well as a deep learning approach.

\textbf{The Random Convolutional Kernel Transform (ROCKET)~\cite{dempster20rocket}} is a transform designed for classification. It generates a large number of parameterised convolutional kernels, used as part of a pipeline alongside a linear classifier. Kernels are randomly initialised with respect to the following parameters: the kernel length; a vector of weights; a bias term added to the result of the convolution operation; the dilation to define the spread of the kernel weights over the input instance; and padding for the input series at the start and end. Each kernel is convoluted with an instance through a sliding window dot-product producing an output vector, extracting only two values: the max value and the proportion of positive values. These are concatenated into a feature vector for all kernels.

\textbf{The Time Series Combination of Heterogeneous and Integrated Embedding Forest (TS-CHIEF)~\cite{shifaz20ts-chief}} is a homogeneous ensemble where hybrid features are embedded in tree nodes rather than modularised through separate classifiers. The trees in the TS-CHIEF ensemble embed distance measures, dictionary based histograms and spectral features. At each node, a number of splitting criteria from each of these representations are considered. These splits use randomly initialised parameters to help maintain diversity in the ensemble.

\textbf{InceptionTime~\cite{fawaz20inception}} is the only deep learning approach we are aware of which  achieves state-of-the-art accuracy for TSC. InceptionTime builds on a residual network (ResNet), the prior best network for TSC~\cite{fawaz19deep}. The network is composed of two blocks of three Inception modules~\cite{szegedy15inception} each, as opposed to the three blocks of three traditional convolutional layers in ResNet. These blocks maintain residual connections, and are followed by global average pooling and softmax layers as before. InceptionTime creates an ensemble of networks with randomly initialised weightings.

\textbf{The Hierarchical Vote Collective of Transform Ensembles, HIVE-COTE 1.0 (HC1)~\cite{bagnall20hivecote1}}, alongside the three algorithms above, are not significantly different to each other in terms of accuracy. Additionally, all are significantly more accurate on average than the best performing algorithms from the bake off comparison of time series classifiers five years prior~\cite{bagnall17bakeoff}.

The second release of HIVE-COTE, \textbf{HIVE-COTE 2.0 (HC2)~\cite{middlehurst21hc2}} is a heterogeneous ensemble of four classifiers built on four different base representations. HC2 is the only algorithm we are aware of which performs significantly better than the four algorithms above. In HC2 three new classifiers are introduced, with only the Shapelet Transform Classifier (STC)~\cite{bostrom15binary} retained from HC1. TDE~\cite{middlehurst20temporal} replaces the Contractable Bag-of-SFA-Symbols (cBOSS)~\cite{middlehurst19scalable}. The Diverse Representation Canonical Interval Forest (DrCIF) replaces both Time Series Forest (TSF)~\cite{deng13forest} and the Random Interval Spectral Ensemble (RISE)~\cite{lines18hive} for the interval and frequency representations. An ensemble of ROCKET~\cite{dempster20rocket} classifiers called the Arsenal is introduced as a new convolutional/shapelet based approach. Estimation of test accuracy via cross-validation is replaced by an adapted form of out-of-bag error, although the final model is still built using all training data.

\subsection{Unsupervised Time Series Transformations}
\label{sec:transforms}

\textbf{Time Series Feature Extraction based on Scalable Hypothesis Tests (TSFresh)~\cite{christ18time}} is a collection of just under 800 features\footnote{\url{https://tsfresh.readthedocs.io/en/latest/text/list\_of\_features.html}} extracted from time series data. TSFresh is very popular with the data science community, and is frequently proposed as a good transform for classification.
The \textbf{Highly Comparative Time Series Analysis (\textit{hctsa})~\cite{fulcher17hctsa}} toolbox can create over 7700 features\footnote{\url{https://hctsa-users.gitbook.io/hctsa-manual/list-of-included-code-files}} for exploratory time series analysis. Alongside basic statistics of time series values, \textit{hctsa} includes features based on linear correlations, trends and entropy. Features from various time series domains such as wavelets, information theory and forecasting among others are also present.
Both TSFresh and \textit{hctsa} cover similar domains, extracting masses of summary features from the time series. Some of these extracted features will be similar, with differently paramaterised variations of the same feature included if applicable.

\textbf{The Canonical Time Series Characteristics (catch22)~\cite{lubba19catch22}} are 22 features chosen to be the most discriminatory of the full \textit{hctsa}~\cite{fulcher17hctsa} set. This was determined by an evaluation over the UCR datasets. The \textit{hctsa} features were initially pruned, removing those which are sensitive to mean and variance and any which could not be calculated on over 80\% of the UCR  datasets. A feature evaluation was then performed based on predictive performance. Any features which performed below a threshold were removed. For the remaining features, a hierarchical clustering was performed on the correlation matrix to remove redundancy.
From each of the 22 clusters formed, a single feature was selected, taking into account balanced accuracy, computational efficiency and interpretability. Like the \textit{hctsa} set it was extracted from, the catch22 features cover a wide range of feature concepts.

\textbf{Time Series Intervals} are used in the interval based representation of TSC algorithms. Classifiers from this representation extract multiple phase-dependent subseries to extract discriminatory features from. Classifiers from this representation include TSF~\cite{deng13forest} and the Canonical Interval Forest (CIF)~\cite{middlehurst20canonical}. Both of these algorithms select intervals with a random length and position, extracting summary features from the resulting subseries and concatenating the output of each. This interval selection and feature extraction process can itself be used as an unsupervised transformation.

\textbf{Generalised Signatures~\cite{morrill20generalised}} are a set of feature extraction techniques primarily for multivariate time series based on rough path theory. We specifically look at the generalised signature method~\cite{morrill20generalised} and the accompanying canonical signature pipeline. Signatures are collections of ordered cross-moments. The pipeline begins by applying two augmentations by default. The basepoint augmentation simply adds a zero at the beginning of the time series, making the signature sensitive to translations of the time series. The time augmentation adds the series timestamps as an extra coordinate to guarantee each signature is unique and obtain information about the parameterisation of the time series. A hierarchical dyadic window is run over the series, with the signature transform being applied to each window. The output for each window is then concatenated into a feature vector.

\section{Experimental structure}
\label{sec:experiments}



We perform our experiments on 112 equal length datasets with no missing values from the UCR time series archive~\cite{dau19ucr}. We resample each dataset randomly 30 times in a stratified manner, with the first resample being the original train-test split from the archive. Each algorithm and dataset resample are seeded using the fold index to ensure reproducibility.

The transformations used in our experiments can be found in the Python \texttt{sktime}\footnote{\url{https://github.com/alan-turing-institute/sktime}} package. Each transformer was built and saved to file, with the process being timed for our timing experiments.
The classification portion of our pipelines, and the TSC algorithms used in our comparison, were run using the Java \texttt{tsml}\footnote{\url{https://github.com/uea-machine-learning/tsml}} toolkit implementations.
An exception for this is the deep learning approach InceptionTime, which we use the \texttt{sktime} companion package \texttt{sktime-dl}\footnote{\url{https://github.com/sktime/sktime-dl}} to run.

To compare our results for multiple classifiers over multiple datasets we use critical difference diagrams~\cite{demsar06comparisons}. We replace the post-hoc Nemenyi test with a comparison of all classifiers using pairwise Wilcoxon signed-rank tests, and cliques formed using the Holm correction as recommended in~\cite{garcia08pairwise,benavoli16pairwise}.

We create pipelines primarily using the transformations described in section~\ref{sec:transforms}, with the exception of the \textit{hctsa} feature set, which required too much processing time and memory to be run in our timeframe.
In addition to these transformations, we also include two benchmark transformations: Principal Component Analysis (PCA) and seven basic summary statistics. The seven statistics we use are the mean, median, standard deviation, minimum, maximum and the quantiles at 25\% and 75\%. PCA and basic summary statistics are the simplest transformations available, and perhaps one of the simplest approaches one could take towards TSC, alongside building classifiers on the raw time series and one-nearest-neighbour classification with Euclidean distance.

Our random interval transformation experiments extract 100 randomly selected intervals per dataset. We form two random interval pipelines, one extracting our basic summary statistics from each interval and the other extracting the Catch22 features.

For the classifier portion of our pipelines, we test three different vector based classifiers. Rotation Forest (RotF)~\cite{rodriguez06rotf} is the classifier of choice for the STC pipeline, and has shown to be significantly better than other popular approaches on problems only containing continuous attributes~\cite{bagnall18rotf}. Extreme Gradient Boosting (XGBoost)~\cite{chen16xgboost} of Kaggle fame is our second classifier option. Our third option is a ridge regression classifiers with cross-validation to select parameters (RidgeCV), the better performing linear classifier suggested for the ROCKET pipeline~\cite{dempster20rocket}.

\section{Results}
\label{sec:results}

We structure our results to answer three specific questions:
\begin{enumerate}
    \item Which transformation is best given a specific classifier?
    \item Which classifier is best, given a specific transform?
    \item How do the pipeline classifiers compare to standard benchmarks?
    \item How do the pipeline classifiers compare to state-of-the-art?
\end{enumerate}

\begin{figure}[htb]
        \centering
        \begin{tabular}{c c}
        \includegraphics[width=0.5\linewidth,trim={2.5cm 5.5cm 1cm 4cm},clip]{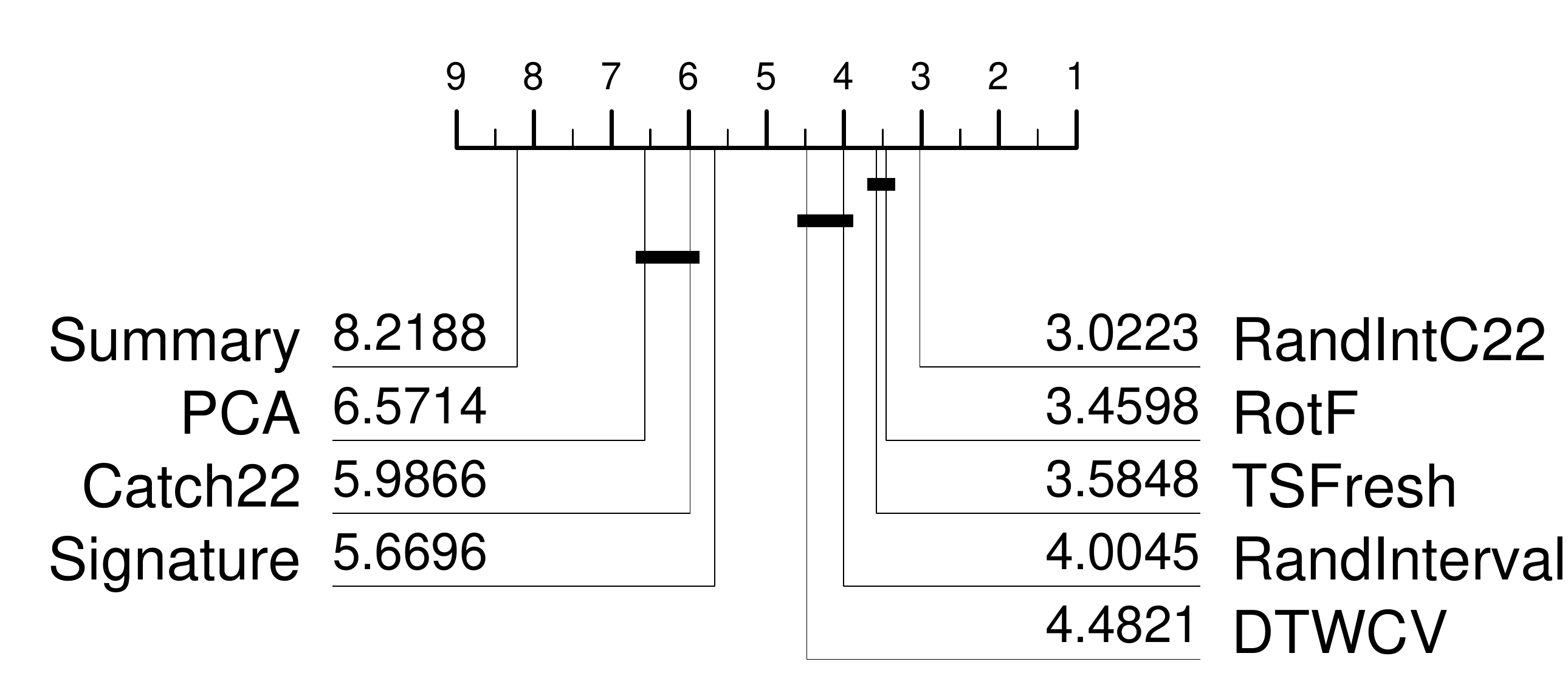} &
        \includegraphics[width=0.5\linewidth,trim={2.5cm 6cm 1cm 4cm},clip]{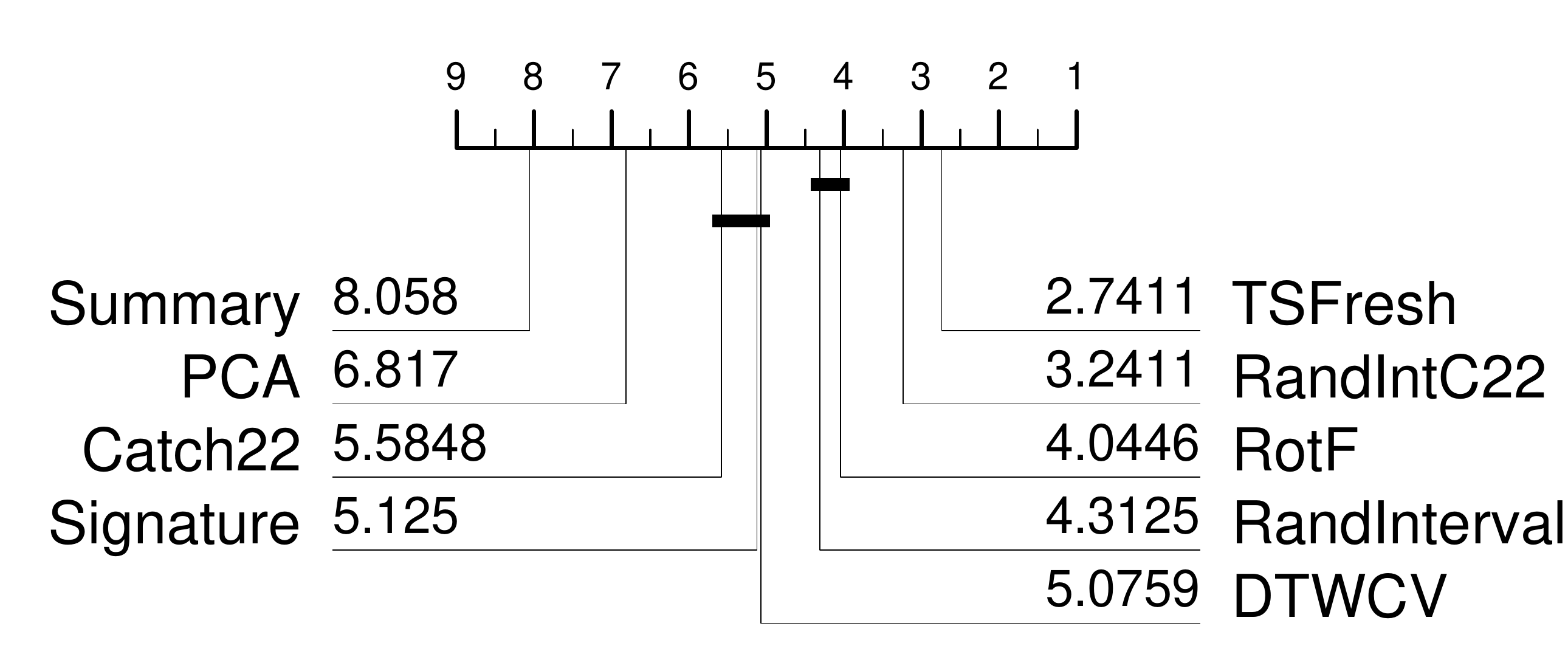}
        \\
        (a) RidgeCV
        &
        (b) XGBoost \\
        \multicolumn{2}{c}{\includegraphics[width=0.5\linewidth,trim={2.5cm 7cm 1cm 4cm},clip]{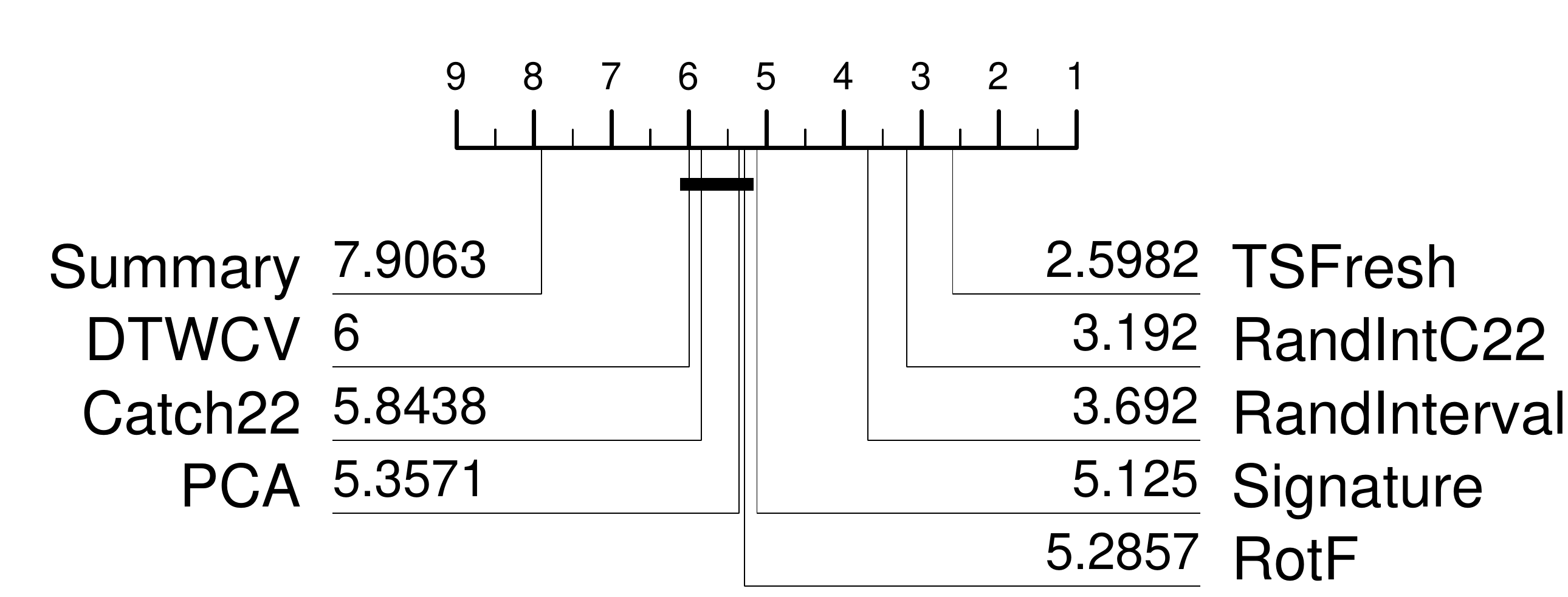}} \\
        \multicolumn{2}{c}{(c) RotF} \\ \\
        \end{tabular}
        \caption{Relative rank performance of seven transforms used in a simple pipeline with a linear ridge classifier (a), XGBoost (b) and rotation forest (c). TSFresh and RandInt22 are significantly better than all other transforms with most base classifiers.}
        \label{tab:transform-cd}
    \end{figure}

Figures~\ref{tab:transform-cd} show the relative performance of difference transforms for our three base classifiers. We include for reference our two baseline classifiers, Rotation Forest (RotF) built on the raw time series and 1-nearest neighbour using dynamic time warping with a tuned window size (DTWCV).


\begin{figure}[b]
    	\centering
        \includegraphics[width=\linewidth,trim={0cm 9cm 0cm 5cm},clip]{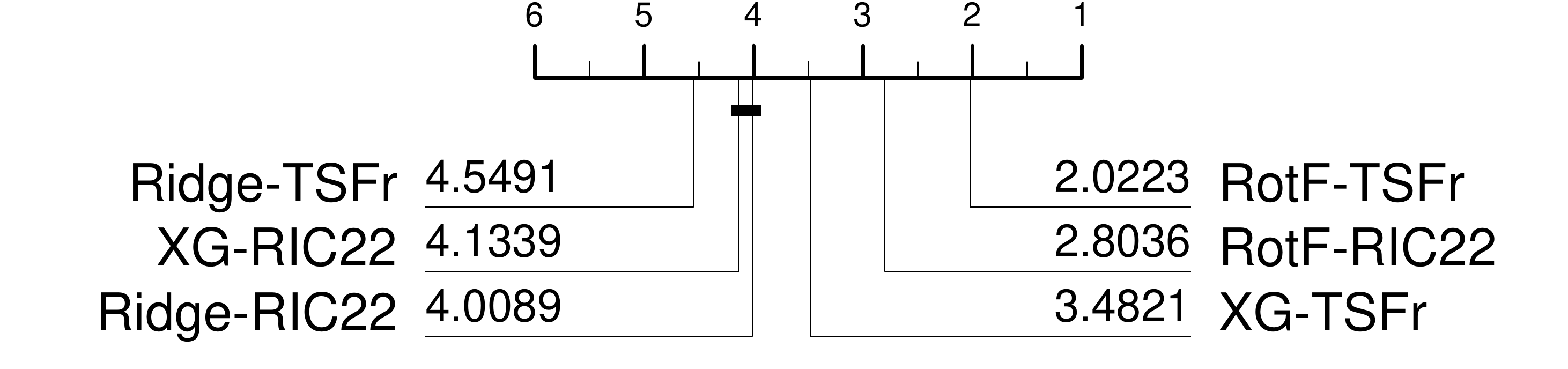}
        \caption{Relative performance of three classifiers Rotation Forest, XBoost and RidgeCV (prefixes RotF, XG and Ridge) with two transforms TSFresh and RandIntCatch22 (suffix TSFr and RIC22). RotF is significantly better than the other classifiers, and RotF with TSFresh is the best overall combination.}
        \label{fig:best-transforms}
    \end{figure}

\begin{figure}[htb]
    \centering
    \begin{tabular}{c c}
        \includegraphics[width=0.5\linewidth,trim={3cm 0cm 0cm 1cm},clip]{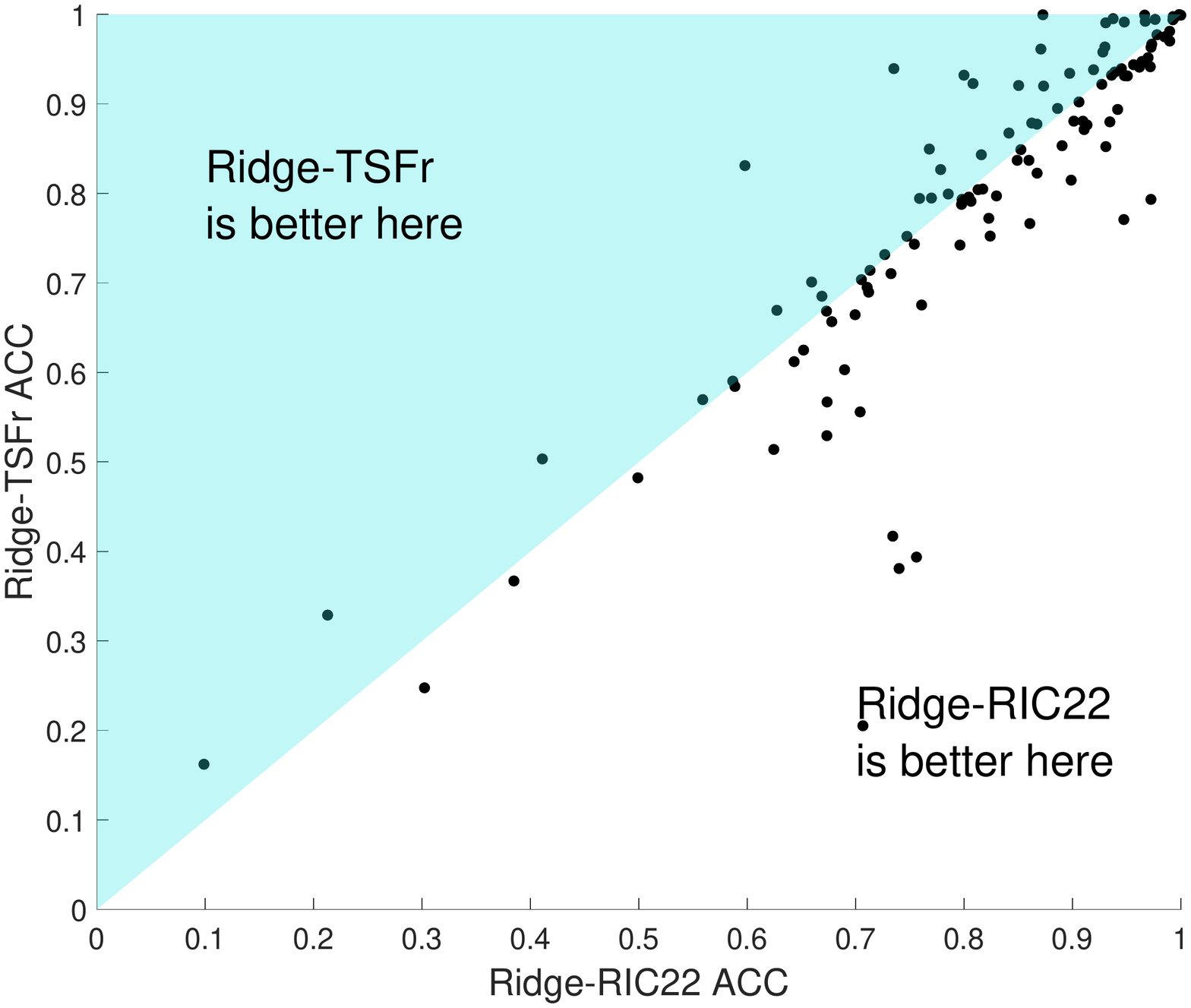} &
        \includegraphics[width=0.5\linewidth,trim={3cm 0cm 0cm 1cm},clip]{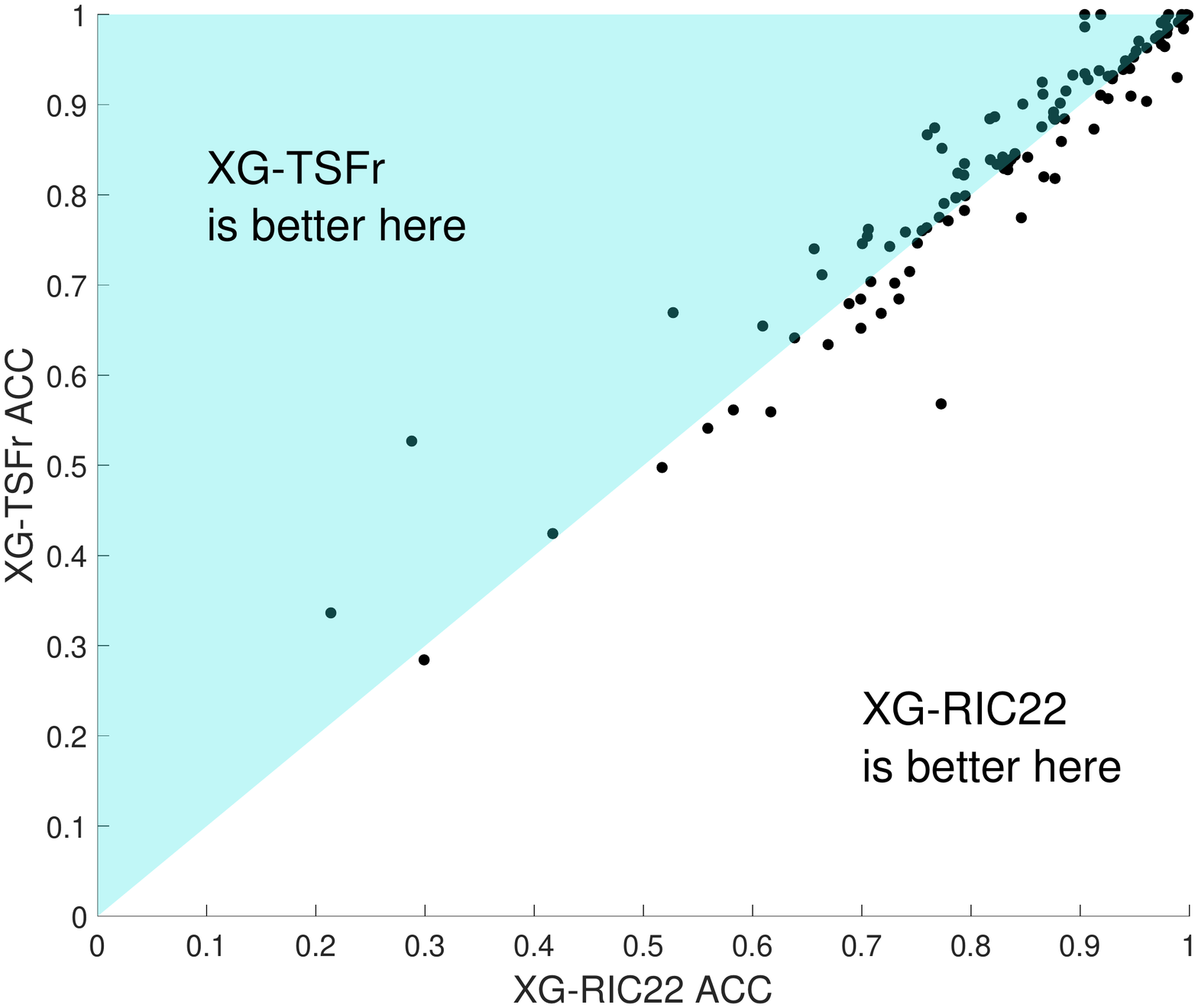}
        \\
        (a) W/D/L: 85/0/27
        &
        (b) W/D/L: 72/0/40 \\

        \includegraphics[width=0.5\linewidth,trim={3cm 0cm 0cm 1cm},clip]{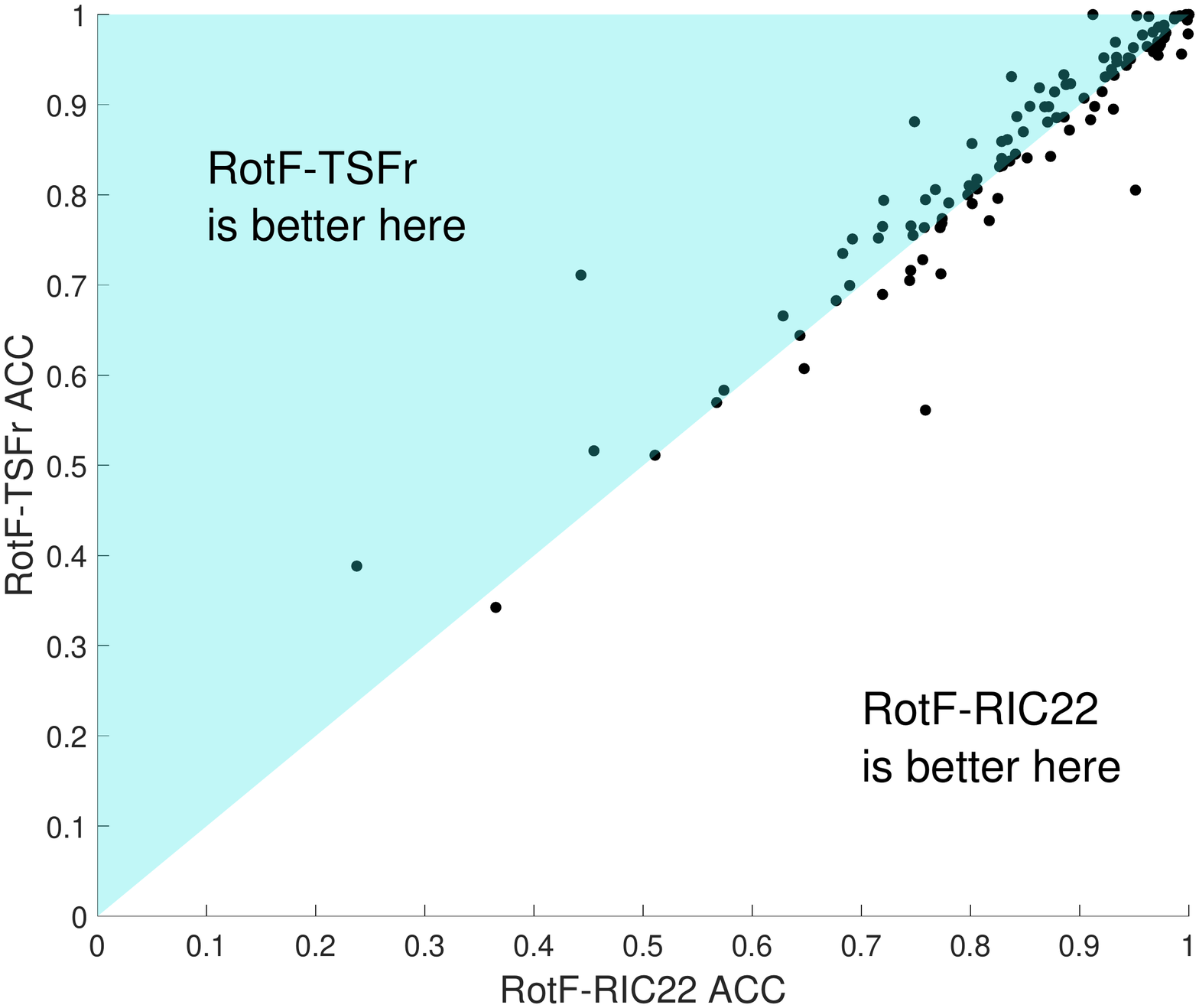}
        &
        \includegraphics[width=0.5\linewidth,trim={3cm 0cm 0cm 1cm},clip]{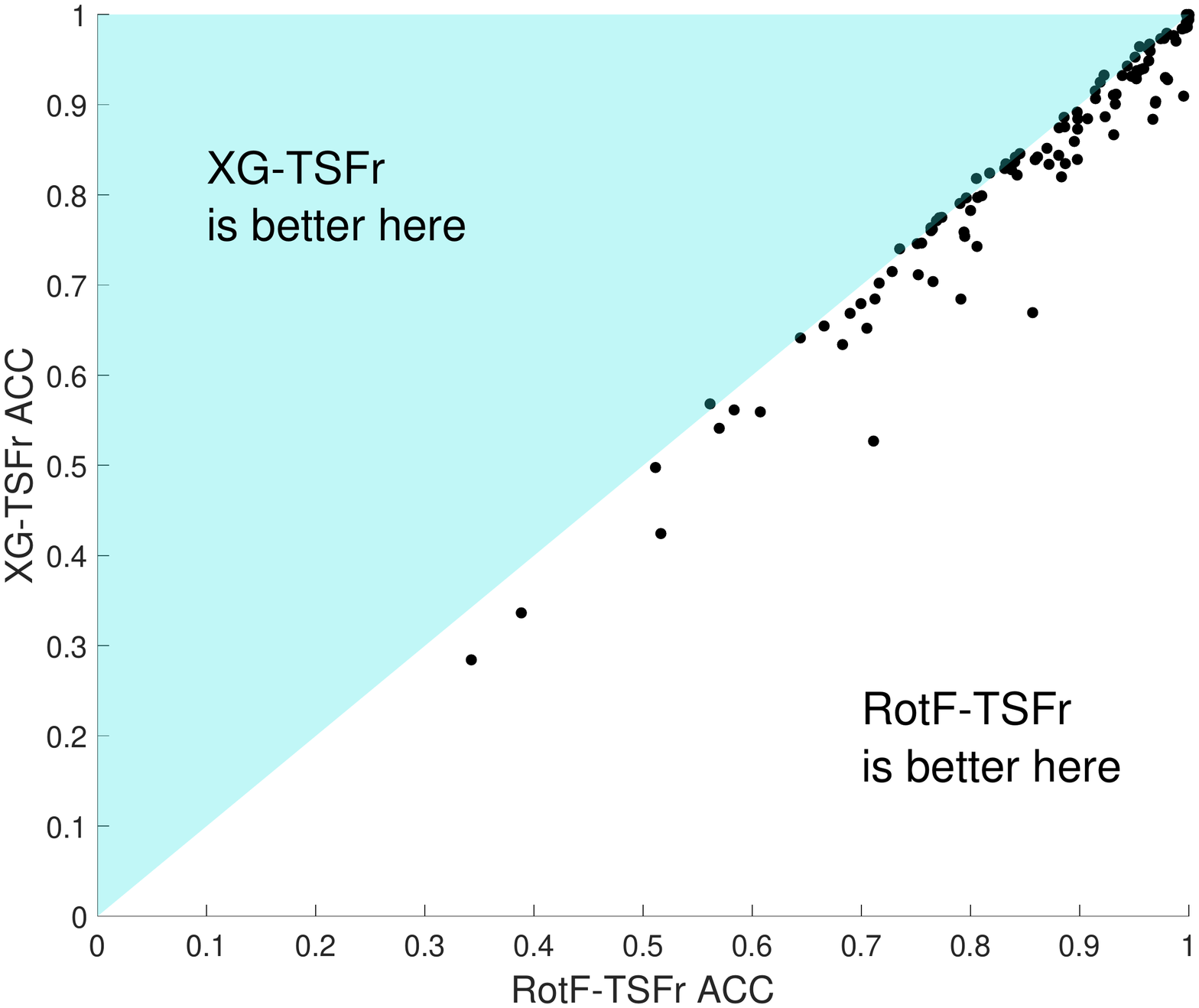}
        \\
        (c) W/D/L: 76/4/32
        &
        (d) W/D/L: 22/3/87 \\ \\
    \end{tabular}
    \caption{Pairwise scatter plots for TSFresh vs RandIntC22 with (a) RidgeCV, (b) XGBoost and (c) rotation forest, and (d) the scatter plot of using TSFresh with XGBoost with TSFresh. (a), (b) and (c) demonstrate the superiority of TSFresh over RandIntC22. (d) shows that rotation forest significantly outperforms XGBoost.}
    \label{tab:pairwise}
\end{figure}

The pattern of results is similar for all three classifiers: TSFresh and RandIntC22 are ranked top for all three classifiers. Both are significantly higher ranked than all the other transforms, and both baseline classifiers, except for the case of TSFresh with a ridge classifier. Summary statistics is always the worst approach and PCA, Catch22, Signatures are no better than, or worse than, the benchmark classifiers. RandomIntervals is significantly better than the benchmarks with rotation forest. There is an anomaly when drawing cliques (RotF and DTW are not always in the same clique despite there being no significant difference in all experiments), but the initial indications are clear: TSFresh and RandIntC22 are the best performing techniques and, with the possible exception of RandomIntervals, the others do not outperform the standard benchmarks, and are therefore of less interest.
We investigate the relative performance of classifiers by comparing the two best transforms (TSFresh and RandIntC22) in combination with the three classifiers. Figure~\ref{fig:best-transforms} shows that RotF is significantly better than RidgeCV and XGBoost for both transforms. This supports the argument made in~\cite{bagnall18rotf} that rotation forest is the best classifier for problems with all continuous attributes.
Figures~\ref{tab:pairwise} show the pairwise scatter plots for four pairs of pipelines. Figures (a), (b) and (c) show the difference in accuracies on the archive for both TSFresh and RandIntC22 using each of our base classifiers. Figure (d) compares our best performing pipeline, TSFresh with rotation forest, to the next best, TSFresh pipeline using XGBoost.

\begin{figure}[htb]
    	\centering
        \includegraphics[width=\linewidth,trim={0cm 9cm 0cm 5cm},clip]{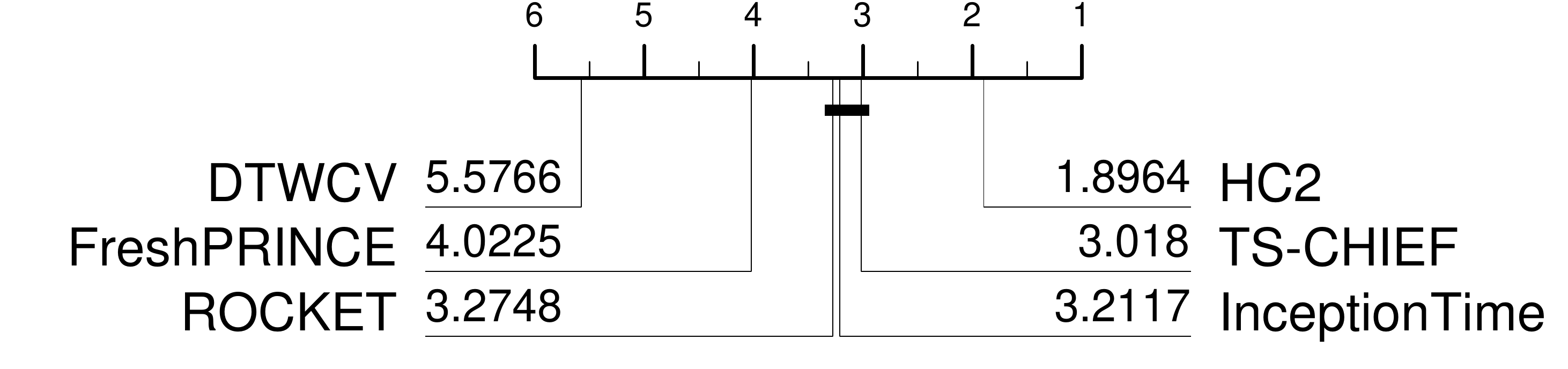}
        \caption{Critical difference plot for FreshPRINCE against SOTA and DTW.}
        \label{fig:sota}
    \end{figure}

Our primary finding is that the pipeline of TSFresh and rotation forest is, on average, the highest ranked and the most accurate simple pipeline approach for classifying data from the UCR archive.
We feel the approach deserves a name better than RotF-TSFr. Hence, we call it the FreshPRINCE (Fresh Pipeline with RotatIoN forest Classifier).
We investigate classification performance of the FreshPRINCE against the current and previous state of the art.  Figure~\ref{fig:sota} shows FreshPRINCE against the very latest state of the art, HIVE-COTEv2.0 (HC2), the previously best performing algorithms, InceptionTime, TS-CHIEF and ROCKET and the popular benchmark, DTWCV.
Figure~\ref{fig:sota} shows that  FreshPRINCE does not achieve SOTA, but it does perform better than the popular benchmark 1-NN with DTW (DTWCV). Table~\ref{tab:summary} presents the summary performance measures averaged over all data. FreshPRINCE is approximately 6.5\% more accurate than DTWCV, but on average 1.4\% and 3.8\% less accurate than ROCKET and HC2.
\begin{table}[htb]
    \centering
        \caption{summary performance statistics averaged over 112 UCR datasets. Test set accuracy (Acc), balanced accuracy (BalAcc), F1 statistic (F1), Area under the receiver operator curve (AUROC) and negative log likelihood (NLL).  }
        \begin{tabular}{l|c|c|c|c|c}\hline
Classifier      &Acc        & BalAcc	&      F1	&    AUROC	& NLL \\ \hline
HC2	            &89.06\%	&	86.85\%	&	0.8575	&	0.9684	&	0.5245 \\
TS-CHIEF	    &87.73\%	&	85.80\%	&	0.8475	&	0.9592	& 0.7479 \\
InceptionTime	&87.36\%	&	85.67\%	&	0.8443	&	0.9583	& 0.6104 \\
ROCKET	        &86.61\%	&	84.58\%	&	0.8339	&	0.9536 	& 1.5754\\
FreshPRINCE	    &85.22\%	&	82.98\%	&	0.8168	&	0.9565  & 0.7230 \\
DTWCV	        &77.72\%	&	76.10\%	&	0.7449	&	0.7860  & 1.4796  \\ \hline
\end{tabular}
    \label{tab:summary}
\end{table}

Table~\ref{tab:timings} displays the run times for generating the results summarised in Figure~\ref{fig:sota} and Table~\ref{tab:summary}. The FreshPRINCE is not as fast as the ROCKET classifier, but is still faster than then other SOTA TSC algorithms.

    \begin{table}[htb]
        \centering
        \caption{RotF, Average (Minutes), Total (Hours), Max (Hours)}
        \label{tab:timings}
            \begin{tabular}{l|c|c|c|c|c|c} \hline
                & DTWCV & TSFresh & Rocket & InceptionTime & TS-CHIEF & HC2 \\ \hline
                Average & 13.9545 & 10.5905 & 1.52939 & 46.3823 & 544.7552 & 182.844 \\
                Total & 26.0485 & 19.7689 & 2.85461 & 86.5802 & 1016.8751 & 341.3084 \\
                Max & 7.3248 & 3.856 & 0.4301 & 7.1093 & 166.7567 & 54.9177 \\ \hline
            \end{tabular}
    \end{table}

We believe that, given the simplicity of the pipeline approach, the FreshPRINCE pipeline should be a benchmark against which new algorithms should be compared. If the claimed merits of an approach are primarily its accuracy, then we believe it should achieve significantly better accuracy than the simple approach of a TSFresh transform followed by a rotation forest classifier.

\subsection{Implementation and Reproduction of Results}

    Given that we suggest FreshPRINCE as a benchmark classifier for new comparisons, we also provide resources for using it as such. We include our results for FreshPRINCE on the 112 UCR datasets used in this experiment on the time series classification web page\footnote{\url{http://www.timeseriesclassification.com/results.php}}. For experiments outside the UCR archive, we have implemented the pipeline in the Python \textit{sktime} package. The most commonly used machine learning package for Python, \textit{sklearn}, does not contain a rotation forest implementation. As such, we also include an implementation of the algorithm in \textit{sktime}.

    Listings~\ref{classifier} displays the process for running FreshPRINCE using the \textit{sktime} package, loading data from its .ts file forma.
\begin{lstlisting}[language=Python, caption={Running the FreshPRINCE pipeline in Python using \textit{sktime}.}, label=classifier]
from sktime.utils.data_io import load_from_tsfile_to_dataframe as load_ts
from sktime.classification.feature_based import FreshPRINCE

if __name__ == "__main__":
    # Load dataset
    trainX, trainY = load_ts("../Data/data_TRAIN.ts")
    testX, testY = load_ts("../Data/data_TEST.ts")

    # Create classifier and build on training data
    fresh_prince = FreshPRINCE()
    fresh_prince.fit(trainX, trainY)

    # Find accuracy on testing data
    accuracy = fresh_prince.score(testX, testY)
\end{lstlisting}
    FreshPRINCE can also be run using a \textit{sklearn} pipeline, using the \textit{sktime} TSFresh transformer and rotation forest implementations as shown in Listing~\ref{pipeline}.
\begin{lstlisting}[language=Python, caption={Forming the FreshPRINCE pipeline in Python using \textit{sktime} components and the \textit{sklearn} Pipeline framework.}, label=pipeline]
    fresh_prince = Pipeline([
            (
                "transform",
                TSFreshFeatureExtractor(
                    default_fc_parameters="comprehensive"),
            ),
            ("classifier", RotationForest()),
        ])
\end{lstlisting}

\section{Conclusion}
\label{sec:conclusions}

We have tested a commonly held belief that a simple pipeline of transformation and standard classifier is a useful approach for time series classification. We have found that there is some merit in this opinion: simple transformations such as PCA or summary stats are not effective, but more complex transformations such as TSFresh and random intervals with the Catch22 features do achieve a respectable level of accuracy on average. They are significantly worse than state of the art in 2021 and 2020, but significantly better than the state from 10 years ago (DTWCV). We suggest the best performing pipeline, a combination of TSFresh and rotation forest we call FreshPRINCE for brevity, be used more commonly as a TSC benchmark.

\section*{Acknowledgements}{
     This work is supported by the UK Engineering and Physical Sciences Research Council (EPSRC) iCASE award T206188 sponsored by British Telecom. The experiments were carried out on the High Performance Computing Cluster supported by the Research and Specialist Computing Support service at the University of East Anglia.
}

\bibliographystyle{splncs03}

\end{document}